\documentclass[11pt]{article}

\usepackage[final]{acl}

\usepackage{times}
\usepackage{latexsym}
\usepackage{enumitem} % Add this to your preamble
\usepackage{subcaption}
\usepackage{amsmath}
\usepackage{url}

\usepackage[T1]{fontenc}
\usepackage[utf8]{inputenc}

\usepackage{microtype}

\usepackage{inconsolata}

\usepackage{graphicx}
\usepackage{booktabs}
\usepackage{tabularx}
\usepackage{multirow}
\newcommand{\simname}{\text{CityBehavEx}}

\newcommand{\parisData}{GreaterParis dataset}
\newcommand{\parisDataShort}{GreaterParis}
\newcommand{\GP}{\text{GP}}
\newcommand{\SH}{\text{SH}}
\newcommand{\cbX}{\text{CBX}}
\newcommand{\AG}{\textit{AG}}
\newcommand{\CS}{\textit{CS}}

\title{
CityBehavEx: A Scalable and Empirically Validated LLM-Assisted Urban Simulation Platform
}

\author{
  \textbf{Gustavo H. Santos\textsuperscript{1,2}}\thanks{\ Corresponding author: \href{mailto:gustavohenriquesantos@alunos.utfpr.edu.br}{gustavohenriquesantos@alunos.utfpr.edu.br}.},
  \textbf{Aline Carneiro Viana\textsuperscript{2}},
  \textbf{Thiago H. Silva\textsuperscript{1,3}}
\\
  \textsuperscript{1}UTFPR, Brazil, \quad
  \textsuperscript{2}Inria, France, \quad
  \textsuperscript{3}University of Toronto, Canada
}

\begin{document}
\maketitle
\begin{abstract}
Recent LLM-based multi-agent urban simulators can generate semantically rich city routines, but they remain costly to scale and are often weakly validated against empirical mobility patterns. We present \simname, an interactive LLM-assisted urban simulation platform that scales to city-size populations, exposes agent behavior for inspection, supports empirical validation, and generates mobility patterns that better match real-world spatial, temporal, and semantic distributions. Instead of invoking large language models for every agent action, \simname{} combines established human mobility models with fine-tuned cross-encoders that estimate semantic alignment between agent profiles, schedules, and activity transitions. This design enables large-scale simulations, as demonstrated in a case study of 100,000 agents over 75 days in under one hour on a single consumer GPU. The platform allows users to define simulation regions, launch experiments, inspect trajectories and activity traces, debug unrealistic behaviors, and validate generated routines against real-world mobility, time-use, and semantic metrics.
\end{abstract}

\section{Introduction}

Recent LLM-based multi-agent urban simulators can generate city routines with semantic richness, sociodemographic awareness, and contextual adaptability \cite{agentsociety, citysim25, ye-etal-2026-mobilecity, wang2024largelanguagemodelsurban}. However, they often rely on repeated LLM inference to generate or revise individual actions, making them costly to scale. Their evaluation is also frequently limited to plausibility checks, LLM-as-a-judge assessments, or coarse aggregate distributions, leaving open questions about whether generated behaviors reproduce empirical patterns of human mobility \cite{santos2026plausiblerealisticevaluatinghuman}. At the same time, human mobility research offers well-established regularities and validation metrics, including daily activity motifs, exploration--return dynamics, visitation patterns, predictability, and spatial displacement laws \cite{Gonzalez2008, Song2010Entropy, Schneider2013, DITRAS, Schlapfer2021-pw}. These findings provide useful building blocks for constraining and evaluating generative urban simulations.

In this paper, we present \simname{}, an interactive LLM-assisted urban simulation platform that scales to city-size populations, exposes agent behavior for inspection, supports empirical validation, and generates mobility patterns that better match real-world spatial, temporal, and semantic distributions. Rather than using LLMs to generate every action of every agent, \simname{} decouples semantic reasoning from trajectory generation. It uses established mobility models to guide daily schedules and exploration--return dynamics, while fine-tuned cross-encoders estimate semantic alignment between agent profiles, schedules, POI choices, and activity transitions. These scores are integrated into efficient stochastic modules for schedule selection, micro-activity generation, social encounters, and transport decisions.

\simname{} is designed as an end-to-end platform for building, inspecting, and validating urban simulations. Accessible via a web interface or CLI, the system enables users to define geographic regions, configure agent populations, launch simulations, and visually replay map trajectories. Post-simulation, users can deeply inspect individual profiles and activity traces, identify unrealistic behaviors, and evaluate generated routines against empirical mobility, time-use, semantic, and social-network metrics. A built-in feedback loop further allows researchers to iteratively adjust parameters and semantic-alignment modules for specific target cities or scenarios.

To ensure continuous public access and prevent closed-source hosting (unlike \textit{CitySim}), the platform is released under the AGPLv3 license\footnote{\url{https://github.com/gefgu/citybehavex}}. The repository includes source code, documentation, a video demonstration, and reproducible configurations. Designed to be lightweight, the default setup runs on a single workstation requiring only a 4GB VRAM GPU for cross-encoder operations, as LLMs can be accessed via API. While larger scenarios are memory-bounded (Appendix \ref{appx:ablation}), we provide a public 1,000-sample YJMOB-based synthetic dataset for out-of-the-box testing without proprietary data.

\simname{} also bridges the gap between scale and realism. A single RTX 5090 simulates 100,000 agents for 75 days in under an hour. For a 500-agent, 7-day scenario, \simname{} takes a few minutes compared to the multi-day runtimes of recent LLM-based baselines. Because it also produces superior spatial, temporal, and semantic mobility patterns, the system shows that efficient language processing can make urban simulation more scalable, inspectable, and empirically realistic.

\section{\simname{} System}

\begin{figure}[!ht]
\centering
\includegraphics[width=0.5\textwidth]{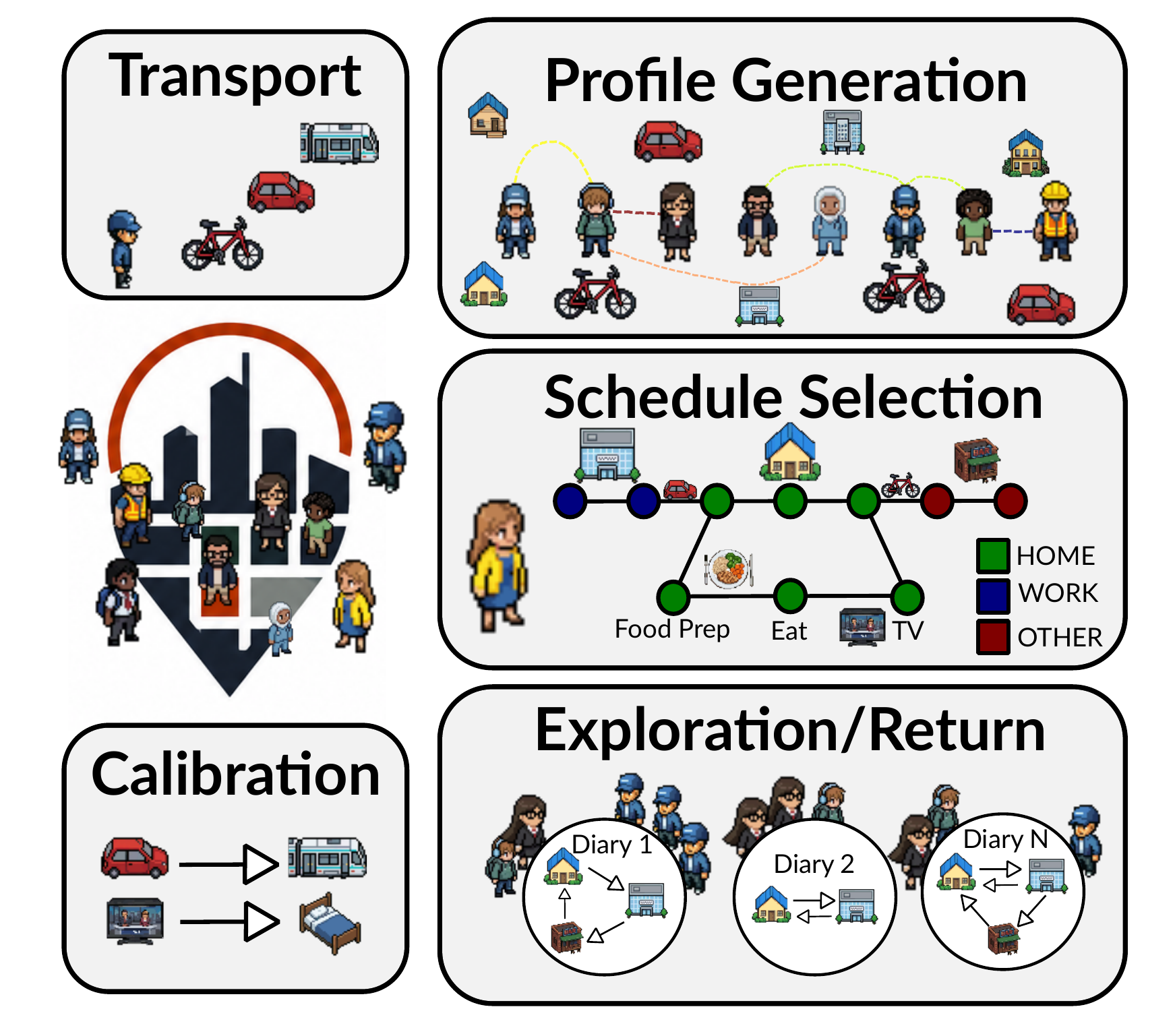}\vspace{-.25cm}
\caption{Overview of the \simname{} architecture.
}
\label{fig:overview}
\end{figure}

This section describes the main modules of \simname{}; Fig.~\ref{fig:overview} summarizes the architecture.

\subsection{Profile Creation}
\label{sec:profile}

Agent profiles provide the demographic and behavioral context used by \simname{} to assign schedules, activities, transport modes, and locations. Profiles are fully configurable, and the system includes a default synthetic population generator for running simulations across cities and scenarios \cite{popSys24}. We adapt the SimPaths profile-generation pipeline \cite{SimPaths}, sampling attributes such as age, education, household composition, occupation, transport resources, and home/work locations; the full schema is provided in Appendix~\ref{app:profile_schema}.

For scalability, the default generator samples attributes independently. To reduce incoherent profiles, such as a 16-year-old agent with a master's degree, a fine-tuned ModernBERT model \cite{modernBERT} flags inconsistent profiles for resampling. Users can also add or redefine attributes by updating the profile schema and the narrative template used by semantic-alignment modules.

Home and work locations receive special treatment because they shape commuting distances and access to nearby opportunities. \simname{} samples home tiles from residential building density and inverse POI density, then samples work tiles from an attractiveness score based on non-residential building density, POI density, and distance decay. By default, home--work distances follow a configurable log-normal distribution estimated from the analyzed datasets (Section~\ref{sec:datasets}). Building and POI data come from Overture Maps \cite{overturemaps2026}.

\subsection{Diary Generation}
\label{sec:methodology_diary}

\paragraph{Generation.}
A diary defines an agent's daily structure as a sequence of activity blocks at 15-minute granularity. \simname{} generates a pool of candidate diaries with an LLM constrained by empirical mobility regularities, including the daily visits law \cite{Schneider2013}. Adapting DITRAS \cite{DITRAS}, locations are represented as \textsc{Home}, \textsc{Work}, or \textsc{Other}. The generator incorporates mobility motifs to produce weekday and weekend routines adapted to the target city, and supports scenario-specific overrides such as holidays, disruptions, or disasters.

\paragraph{Semantic alignment and clustering.}
Recognizing that human mobility largely follows a limited set of predictable daily motifs \cite{Schneider2013}, \simname{} avoids the prohibitive cost of per-agent LLM inference during diary assignment. Instead, we estimate semantic compatibility between textualized agent profiles and candidate diaries using a fine-tuned ModernBERT cross-encoder \cite{modernBERT}. To train this model, we prompt an LLM to generate 1,000 diverse profile–diary pairs scored for compatibility. These pairs serve as the supervision signal, fine-tuning the cross-encoder to map new profile–diary combinations to an alignment score $s_k$. To further ensure performance scaling, agents are clustered by profile similarity, and alignment scores are computed at the cluster level. Finally, stochastic assignment based on these scores introduces natural behavioral variation among agents within the same cluster.

\paragraph{Semantically weighted CRP.}
To balance routine with exploration, \simname{} uses a semantically weighted Chinese Restaurant Process (SW-CRP) \cite{ddcrp}. Simulation days are treated as ``customers'' and candidate diaries as ``tables.'' Let $K$ denote the set of diaries already selected by the agent in previous days, and $U$ the set of candidate diaries not yet used by that agent. The probability that an agent selects an already-used diary $k \in K$ on day $t$ depends on both prior usage and semantic alignment:
$$
P(c_t = k) =
\frac{n_k \cdot e^{s_k/T}}
{\sum_{j \in K} n_j \cdot e^{s_j/T} + \sum_{j \in U} \alpha \cdot e^{s_j/T}},
$$
\noindent where $n_k$ is the usage count for diary $k$, $s_k$ is the semantic alignment score, $T$ is a temperature parameter, and $\alpha$ controls exploration of unused diaries in $U$. For an unused diary $k \in U$, the same denominator is used with numerator $\alpha e^{s_k/T}$. To model heterogeneity, $T$ and $\alpha$ are sampled from configurable log-normal distributions.

\subsection{Exploration and Preferential Return}

During diary execution, agents travel to fixed \textsc{Home} and \textsc{Work} tiles, while destinations for \textsc{Other} blocks are selected through an Exploration and Preferential Return (EPR) mechanism \cite{song2010modellingEPR}. Agent-specific parameters determine whether an agent explores a new POI or returns to a previously visited location. To make exploration semantically aware, \simname{} scores POI types with a fine-tuned ModernBERT model and applies the SW-CRP mechanism described in Section~\ref{sec:methodology_diary}. For each profile cluster and \textsc{Other} block, this process filters POI categories that are compatible with the agent and context. Exploration then samples a new destination from the filtered POI types, while return samples from the agent's visited locations.

\subsection{Micro-Schedule Generation}

Recent LLM-based urban simulators often generate open-vocabulary micro-activities, such as \textit{reading a book} or \textit{writing an email} \cite{agentsociety}. Although expressive, such descriptions are difficult to validate against time-use data. \simname{} instead grounds micro-schedule generation in the 25 activity classes defined by the Multinational Time Use Study (MTUS) \cite{ipums_mtus}.

When an agent enters a macro-schedule block, such as \textsc{Home}, \textsc{Work}, or \textsc{Other}, \simname{} samples activities until the block ends. We contextually mask the activity set: sleeping is disabled during \textsc{Work} blocks, while \textsc{Other} activities are restricted by POI types mapped from Overture Maps into 15 categories using an LLM. To vary block transitions, we sample activity durations from activity-specific log-normal distributions. This provides a simple, controllable distribution that is easier for LLMs to adapt, building on classic models \cite{moore1997generation,strum2000modeling}.

Activity choices use the same semantically weighted CRP mechanism as diary selection. The semantic weights are conditioned on the profile cluster (Section~\ref{sec:methodology_diary}), day period, block or POI type, and previous activity. This lets \simname{} encode temporal common sense: an agent arriving home at midnight is more likely to sleep, whereas an evening home block assigns higher probability to preparing and eating food.

\subsection{Social Module}

\simname{} models social network formation through spatial proximity, semantic similarity, and co-location. The initial network samples each agent's expected number of friends from a configurable log-normal degree distribution, with candidates drawn from the agent's home and work H3 cells. Higher profile-embedding similarity increases the probability of forming a tie. During the simulation, casual friendships emerge from repeated co-location: agents who visit the same places may connect based on encounter regularity and neighborhood overlap, following RECAST \cite{recast2015}. Friendship strength is updated at configurable intervals, increasing after repeated encounters and decaying when no encounters occur.

\subsection{Transportation Module}

\simname{} uses a multimodal transportation module to assign transportation modes and estimate movement between locations. Each agent has walking and cycling distance thresholds sampled from log-normal distributions. Trips below the walking threshold are assigned to walking; otherwise, mode choice depends on the agent's transport resources from profile creation (Section~\ref{sec:profile}). Agents with bicycle access cycle when the trip is within their cycling threshold, agents with car access use the road network, and remaining trips use rail when available or car/taxi travel as a fallback.

Road and rail are obtained from Overture Maps \cite{overturemaps2026}, including speed-limit information when available, constraining routes and travel times to real infrastructure. To keep routing efficient, \simname{} uses a cacheable contraction-hierarchies method \cite{contractionHierachies}, which computes shortest paths and reuses routes across agents and repeated trips.

\subsection{Feedback and Calibration Loop}
\label{sec:feedback_loop}

Because \simname{} relies on configurable parameters and semantic-alignment models, users can iteratively calibrate simulations after inspecting the outputs. The feedback loop supports two validation settings: comparing generated behavior against mobility laws from the literature, or, when empirical mobility or time-use data are available, comparing simulations against the full evaluation suite.

\simname{} supports both manual and automated calibration. Through the web interface, users can inspect trajectories, diagnose unrealistic behaviors, and adjust parameters. Through the CLI, users or coding agents can launch simulations, retrieve evaluation results, modify configuration files or semantic-alignment prompts, and rerun experiments. Optionally, users can configure a fixed number of automated update rounds, in which the simulation LLM proposes parameter or prompt updates. This process helps adapt simulations to specific cities, cultures, or special-event scenarios while keeping calibration inspectable and reproducible.

\subsection{User Interface}
\label{sec:UI}
 
\simname{} provides a web interface for configuring, running, inspecting, and validating urban behavior simulations. Users can select a geographic bounding box, configure the agent population, run pre-simulation steps, launch the full simulation, and replay trajectories on a Mapbox basemap. During replay, users can inspect each agent's profile, macro-schedule, micro-activities, transport choices, and social graph, allowing unexpected behaviors to be traced back to specific simulation components.

The validation dashboard, illustrated in Appendix~\ref{app:valDash}, supports interactive comparison across simulation runs, empirical datasets, and mobility laws from the literature. It includes mobility, temporal, topological, behavioral, semantic, and social metrics, with the full list in Appendix~\ref{app:metrics_table}. Metrics can be filtered by day type, such as weekdays, weekends, and special days, and by day period.

\begin{figure}[!ht]
\centering
\includegraphics[width=0.5\textwidth]{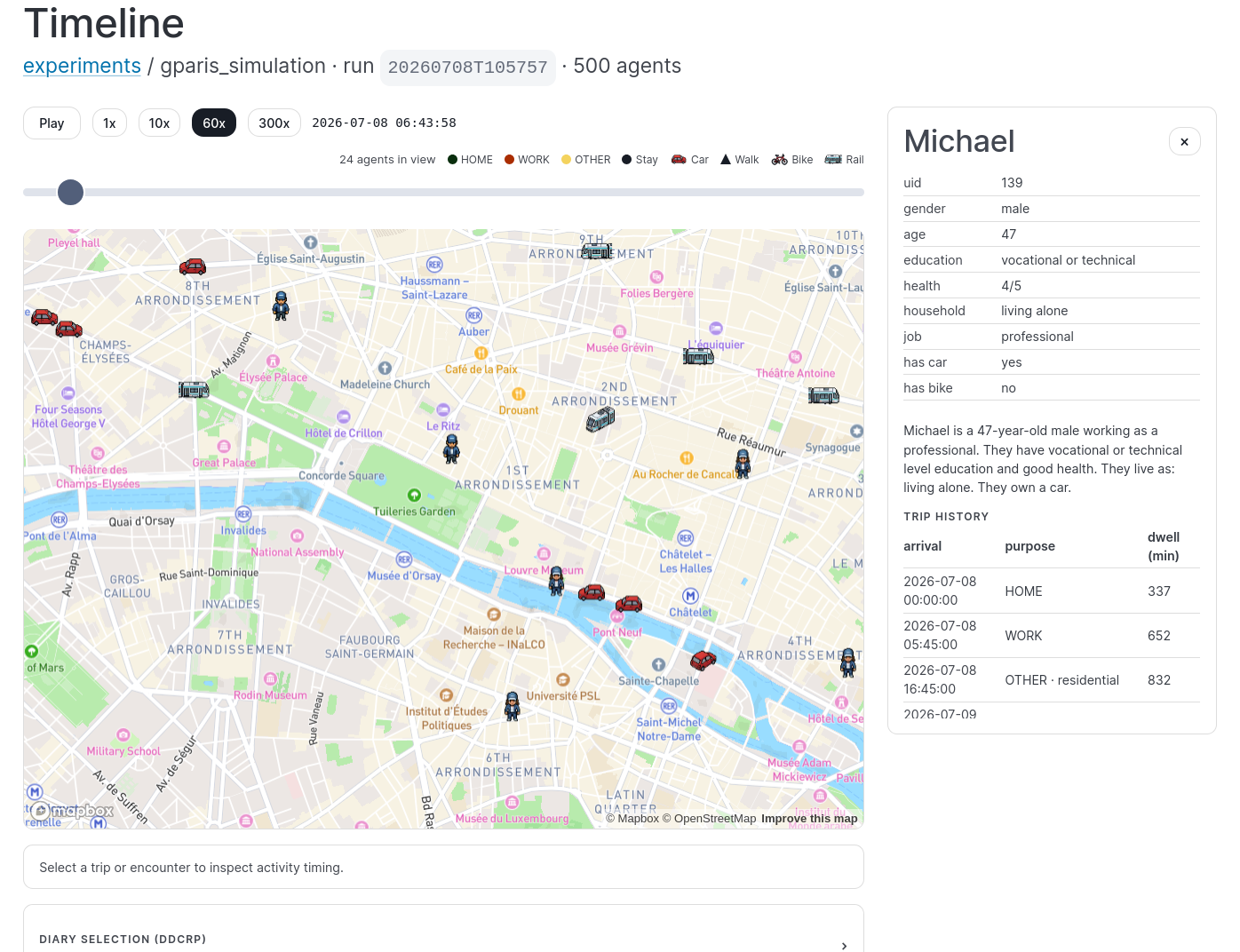}\vspace{-.25cm}
\caption{\textbf{Timeline View}. This interactive interface allows users to inspect the ongoing simulation and replay agent trajectories. Through the side panel, users can examine detailed agent profiles, daily schedules, and micro-activities. Additionally, the map utilizes AI-generated sprites to ease the visualization of agents, their states, and transportation modes across the city.}
\label{fig:timeline}
\end{figure}

To keep the dashboard responsive, the visual analytics use a custom Python--Rust implementation of Scikit-Mobility functionality \cite{ScikitMobility}, computing the evaluation suite in seconds.

Simulated distributions are compared against empirical data using metric-specific distances. Jensen--Shannon Divergence (JSD) is used for categorical distributions, including visit purposes, activity transitions, daily routines, spatio-temporal visits, motifs, and profile distributions. The Common Part of Commuters (CPC)~\cite{ScikitMobility} measures overlap between empirical and simulated origin--destination matrices. For numerical or ordinal distributions, such as travel distance, trip duration, radius of gyration, dwell time, and visitation frequency, \simname{} uses Wasserstein distance ($W_1$). All metric values can be visualized in the dashboard and exported for offline analysis.

In addition to the web interface, \simname{} provides a CLI for batch execution, automated calibration, and integration with coding agents that can run simulations, retrieve evaluation results, and modify configuration files.

\section{Evaluation}

\subsection{Datasets}
\label{sec:datasets}

We evaluate \simname{} on three real-world mobility datasets that vary in scale, duration, spatial resolution, and semantic annotation:

\begin{itemize}[noitemsep, topsep=2pt, leftmargin=*]
\item \textbf{\parisData{}}~\cite{chasse:hal-05365088}: A non-public 7-day anonymized GNSS dataset from the Île-de-France region, with trajectories for 3,337 individuals mapped to H3 level-10 cells and annotated with trip purpose and transportation mode.

\item \textbf{Shanghai}~\cite{esper_sig24}: A non-public 10-day CDR dataset with hourly movements of 58,502 users across 10,396 grid cells of 200m $\times$ 200m, providing broad population coverage but limited semantic annotation.

\item \textbf{YJMOB}~\cite{YJMob100K}: A public smartphone-location dataset tracking 100,000 users in Nagoya, Japan, over 75 days at 30-minute granularity across 40,000 grid cells of 500m $\times$ 500m. We also use its 25,000-user disaster subset, with 60 regular days followed by 15 disaster days.
\end{itemize}

\subsection{Simulator Comparison}
\label{sec:simulator_comparison}

We compare \simname{} (\cbX{}) with two recent LLM-based urban simulators, \textit{AgentSociety} (\AG{}) and \textit{CitySim} (\CS{}), using spatial, temporal, and semantic realism metrics computed against the datasets in Section~\ref{sec:datasets}. The comparison follows prior work showing that plausible LLM-generated narratives can still diverge from observed mobility patterns \cite{santos2026plausiblerealisticevaluatinghuman}. Additional full-scale results and ablations are provided in Appendix~\ref{appx:ablation}.

\subsubsection{Mobility Realism}

\begin{table*}[t]
\centering
\scriptsize
\setlength{\tabcolsep}{3pt}
\caption{Spatial, temporal, and semantic realism of simulated trajectories. Lower $W_1$, VPD, ATM, and DARD values indicate closer agreement with empirical data. GP=GreaterParis, SH=Shanghai, AG=AgentSociety, CS=CitySim, and CBX=\simname{}. Semantic metrics are reported only for datasets with purpose labels.}
\label{tab:comprehensive_results_ag_comparision}
\begin{tabular}{ll ccc ccc ccc}
\toprule
\textbf{Dataset} & \textbf{Source} & \textbf{\textit{$\Delta r$}} ($W_1$) & \textbf{\textit{$r_g$}} ($W_1$) & \textbf{\textit{TD} (min)} ($W_1$) & \textbf{\textit{DT} (h)} ($W_1$) & \textbf{\textit{Vf}} ($W_1$) & \textbf{\textit{VPD}} ($10^{-2}$) & \textbf{\textit{ATM}} ($10^{-2}$) & \textbf{\textit{DARD}} ($10^{-2}$) \\
\midrule

\GP{} & \AG{} & $14.8 \pm 0.4$ & $7.3 \pm 0.3$ & $24.4 \pm 0.2$ & $4.0 \pm 0.1$ & $9.6 \pm 0.2$ & $3.0 \pm 0.1$ & $11.3 \pm 0.3$ & $\mathbf{1.1 \pm 0.0}$ \\
\GP{} & \CS{} & $7.5 \pm 1.7$ & $3.5 \pm 0.2$ & $24.3 \pm 18.9$ & $30.2 \pm 29.6$ & $19.5 \pm 3.6$ & $8.8 \pm 2.6$ & $24.1 \pm 5.6$ & $15.7 \pm 21.6$ \\
\textbf{\GP{}} & \textbf{\cbX{}} & $\mathbf{1.32 \pm 0.23}$ & $\mathbf{0.93 \pm 0.16}$ & $\mathbf{5.33 \pm 0.45}$ & $\mathbf{1.15 \pm 0.01}$ & $\mathbf{4.09 \pm 0.04}$ & $\mathbf{0.10 \pm 0.02}$ & $\mathbf{7.16 \pm 0.42}$ & $3.69 \pm 0.79$ \\

\midrule

\SH{} & \AG{} & $8.7 \pm 0.5$ & $4.3 \pm 0.7$ & $9.3 \pm 0.2$ & $4.8 \pm 0.4$ & $\mathbf{12.3 \pm 4.2}$ & -- & -- & -- \\
\SH{} & \CS{} & $4.0 \pm 0.1$ & $4.9 \pm 0.1$ & $19.3 \pm 0.3$ & $36.4 \pm 4.9$ & $25.0 \pm 0.7$ & -- & -- & -- \\
\textbf{\SH{}} & \textbf{\cbX{}} & $\mathbf{2.03 \pm 0.01}$ & $\mathbf{1.48 \pm 0.02}$ & $\mathbf{4.96 \pm 0.05}$ & $\mathbf{2.56 \pm 0.01}$ & $13.87 \pm 0.04$ & -- & -- & -- \\

\midrule
\SH{} & Ref. & $0.5$ & $1.0$ & $0.3$ & $0.3$ & $2.1$ & -- & -- & -- \\

\bottomrule
\end{tabular}
\end{table*}

Table~\ref{tab:comprehensive_results_ag_comparision} reports results for 500-agent samples averaged over three runs; larger direct comparisons were infeasible because each baseline run required approximately two to three days. The ``Ref.'' row compares two disjoint empirical samples from the same dataset, providing a lower-bound reference for the expected distance between real mobility samples. Across both datasets, \simname{} obtains lower distances for spatial metrics---travel distance ($\Delta r$) and radius of gyration ($r_g$)---and temporal metrics, including trip duration (TD), dwell time (DT), and visitation frequency (Vf). For semantic realism, we report Visit Purpose Distribution (VPD), Activity Transition Matrix (ATM), and Daily Activity Routine Distribution (DARD) only for \parisDataShort{}, which provides trip-purpose labels. Across both datasets, CityBehavEx achieves the lowest distances for all evaluated spatial and temporal metrics except visitation frequency in Shanghai, for which AgentSociety obtains a slightly lower distance (12.3 versus 13.87). For the GreaterParis semantic metrics, CityBehavEx achieves the lowest VPD and ATM distances, while AgentSociety achieves the lowest DARD distance.

\subsubsection{Runtime and Scalability}

For a 500-agent, 7-day scenario, \simname{} completes execution in $2.49 \pm 1.7$ minutes, while \textit{AgentSociety} and \textit{CitySim} require $78.28 \pm 8.21$ and $83.9 \pm 4.50$ hours, respectively. These runtimes exclude map-building caches, which are also required by the baselines, and model fine-tuning, which takes approximately 5--20 minutes per model. \simname{} also simulates 100,000 agents over 75 days in $29.2 \pm 0.6$ minutes on YJMOB, demonstrating scalability beyond the small scenarios feasible for direct comparison. LLMs and ModernBERT calls can be cached independently, further speeding up iterative calibration, which makes the same scenario run in less than 10 minutes.

Together, these results show that \simname{} supports interactive experimentation while producing mobility patterns that more closely match empirical data than recent LLM urban simulators.

%\subsection{Simulator advantages}
\subsection{Scenario and Module-Level Evaluation}
\label{sec:evaluation_modules}

Beyond direct simulator comparison, we evaluate \simname{} on capabilities that are explicit in our system but absent, implicit, or not evaluated in the baselines: social network formation, special-event adaptation, transport choice, and time-use realism.

\paragraph{Social network realism.}
To distinguish between random encounters and true friendships within the Shanghai and YJMOB mobility datasets, we apply the RECAST framework \cite{recast2015}. We evaluate how well our simulated networks mirror real-world inferred social networks by comparing the distributions of degree, clustering coefficients, edge regularity, and topological overlap, quantifying the similarity using the Wasserstein ($W_1$) distance (full results in Appendix \ref{appx:ablation}). Focusing specifically on the clustering coefficient, the distributions produced by \simname{} closely align with real-world data ($W_1 = 0.0284$). This substantially outperforms the comparison with the degree-preserving random network baseline generated by the RECAST framework ($W_1 = 0.2211$). Overall, this demonstrates that \simname{} agents successfully cluster and engage in joint activities in a manner that strongly reflects real-world human behavior.

\paragraph{Disaster scenario.}
We evaluate special-event adaptation using the YJMOB disaster subset, which contains 60 regular days and 15 disaster days. We compare pre/during-event mobility using travel distance, visitation frequency, radius of gyration, activity routines, and spatio-temporal visit distributions. The validation interface lets users inspect these changes separately for regular and special-event days. As shown in more detail in Appendix \ref{appx:ablation}, the mobility metrics still show strong alignment even under a disaster scenario.

\paragraph{Transport mode choices.}
We study transport mode choice using \parisDataShort{}, the only one with mode annotations. Compared with the empirical mode distribution, \simname{} achieves a JSD of $0.039 \pm 0.0002$. The main discrepancy is car versus rail usage: agents use cars in $58.37 \pm 1.034\%$ of trips, compared with $36.9\%$ in \parisDataShort{}.

\paragraph{Time-use realism.}
We evaluate micro-schedules against MTUS 2009 time-use distributions for France. \simname{} achieves a low mean absolute day-share difference of $2.98 \pm 0.009$, with the largest discrepancies at night, where agents sleep less and engage more in other activities than observed in MTUS.

\section{Limitations}

\simname{} has several limitations. First, reproducibility is constrained by dataset access: among the datasets used in our evaluation, only YJMOB is publicly available. Second, because \simname{} uses LLMs and fine-tuned language models for semantic alignment and calibration, it may inherit cultural, gender, and socioeconomic biases from these models or from the data used to fine-tune them. Third, our evaluation indicates that spatial allocation remains a challenge. In particular, STVD and CPC results suggest that agents may choose semantically appropriate activities and POIs, but allocate them to neighborhoods that differ from those observed in empirical data. Future work will incorporate neighborhood-level characteristics more directly into mobility and POI choice.

\section{Conclusion}

We presented \simname{}, an interactive LLM-assisted platform for scalable and empirically validated urban behavior simulation. By combining mobility models, fine-tuned semantic alignment, and a parallelized Rust core, \simname{} avoids costly per-action LLM inference while preserving context-aware behavioral variation. Our evaluation shows that \simname{} runs orders of magnitude faster than recent LLM-based urban simulators and produces trajectories that more closely match empirical spatial, temporal, and semantic mobility distributions. The web interface further supports inspection, validation, and iterative calibration, making \simname{} a practical tool for realistic city-scale mobility generation.

\section*{Acknowledgments}
This study was supported by CNPq (processes 314603/2023-9, 441444/2023-7, 444724/2024-9, and INCT TILD-IAR 408490/2024-1) - and the PEPR MOBIDEC Mob Sci-Dat Factory project. 

\clearpage

\section*{Ethics Statement}

This paper introduces CityBehavEx, a scalable and inspectable platform for language-model-assisted urban behavior simulation. While our framework offers significant advantages for reproducing empirically grounded human mobility and studying urban dynamics at scale, it raises important ethical considerations that must be responsibly addressed.

A primary concern is the potential for bias amplification. CityBehavEx relies on default synthetic population generators that sample attributes such as age, gender, education, and occupation, alongside fine-tuned cross-encoders and LLM constraints for semantic alignment. Any societal biases inherent in the foundation language models or the empirical mobility training datasets may propagate through the simulation. This risks generating skewed or stereotypical representations of population behaviors. If these simulated outcomes are used to inform real-world urban policies or infrastructure design, they could inadvertently reinforce existing socioeconomic or demographic inequities.

Furthermore, the platform's capacity to model highly realistic, context-aware trajectories, transportation choices, and dynamic social network formation introduces risks regarding the potential misuse of simulation insights. Predicting detailed mobility trends and individual routines could theoretically be leveraged for unethical surveillance, behavioral manipulation, or commercial exploitation without public consent. Additionally, while the system's evaluation utilizes anonymized real-world mobility records (such as the Greater Paris, Shanghai, and YJMOB datasets), generating semantically rich, city-scale routines necessitates ongoing safeguards to ensure synthetic behaviors do not expose or target vulnerable real-world communities.

To mitigate these risks, CityBehavEx is intentionally designed with an interactive feedback loop and a visual validation dashboard, explicitly supporting human oversight, inspectability, and debugging of unrealistic behaviors. We strongly advocate that synthetic urban agents should be employed to complement, rather than replace, the essential involvement of actual residents, domain experts, and stakeholders in urban planning processes. By prioritizing transparency and iterative human validation, we aim to ensure the responsible and ethical deployment of generative agents in social and urban studies.

% Bibliography entries for the entire Anthology, followed by custom entries
%\bibliography{anthology,custom}
% Custom bibliography entries only

\clearpage

\appendix

\begin{table*}[!htbp]
\centering
\scriptsize
\caption{Metrics and empirical laws supported by the \simname{} validation dashboard.}\vspace{-0.4cm}
\label{tab:metrics_available}
\begin{tabular}{p{0.16\textwidth}p{0.79\textwidth}}
\toprule
\textbf{Type} & \textbf{Metrics} \\
\midrule
Spatial laws and metrics & \textbf{Travel distance ($\Delta r$)}~\cite{Gonzalez2008}, \textbf{Radius of Gyration ($r_g$)}~\cite{Gonzalez2008}, \textbf{Daily visits}~\cite{Schneider2013}, \textbf{Predictability}~\cite{Song2010Entropy}, \textbf{Distance-frequency}~\cite{Schlapfer2021-pw}, OD matrix~\cite{ScikitMobility} \\
Temporal metrics & Trip duration (TD), Dwell time (DT), Visitation frequency (Vf)~\cite{Schneider2013} \\
Topological laws & \textbf{Mobility motifs}~\cite{Schneider2013} \\
Behavioral metrics & Mobility profiles~\cite{licia2021}, Regularity, Stationarity and Diversity~\cite{Teixeira2021}, Entropy~\cite{Teixeira2021} \\
Semantic metrics & Visit Purpose Distribution (VPD), Activity Transition Matrix (ATM)~\cite{senefonteSocInfo2020}, Daily Activity Routine Distribution (DARD)~\cite{wang2024largelanguagemodelsurban}, Spatio-Temporal Visit Distribution (STVD)~\cite{wang2024largelanguagemodelsurban} \\
Social Metrics & Node Degree, Clustering Coefficient, Edge Regularity and Topological Overlap \cite{recast2015}  \\
Time-Use Metrics & Mean Absolute day-share difference \\
Transport Metrics & Transport Choice Distribution (TCD) \\
\bottomrule
\end{tabular}
\end{table*}

\begin{table*}[!htbp]
\centering
\scriptsize
\caption{Comprehensive spatial and temporal mobility realism metrics, including an ablation study. The evaluation uses four datasets: \parisData{} (1,500 agents, 7 days), Shanghai (29,251 agents, 10 days), YJMOB (50,000 agents, 75 days), and YJMOB disaster (25,000 agents, 75 days: 60 normal, 15 disaster). The ablation columns demonstrate the impact of independently removing the Profile, Micro-schedule, Social, and Transport modules. Dataset populations are split equally between the reference sample and the \simname{} simulation. \textit{RT} is runtime. Ref.: distance between two disjoint, equally sized empirical samples from the same dataset. }
\label{tab:combined_results_vertical_ablation}
\begin{tabular}{p{1.2cm}lcccccc}
\toprule
\textbf{Dataset} & \textbf{Metric ($W_1$)} & \textbf{\simname} & \textbf{- Profile} & \textbf{- Micro-sched.} & \textbf{- Social} & \textbf{- Transport} & \textbf{Ref.} \\
\midrule
\parisData{} & \textit{$\Delta r$} & $\mathbf{1.0 \pm 0.0}$ & $4.0 \pm 0.0$ & $1.9 \pm 0.2$ & $1.1 \pm 0.1$ & $1.0 \pm 0.1$ & $0.2$ \\
 & \textit{$r_g$} & $\mathbf{0.6 \pm 0.1}$ & $1.2 \pm 0.1$ & $2.4 \pm 0.2$ & $\mathbf{0.6 \pm 0.1}$ & $0.6 \pm 0.2$ & $0.2$ \\
 & \textit{TD.} (min) & $4.8 \pm 0.4$ & $4.8 \pm 0.1$ & $2.2 \pm 0.2$ & $4.6 \pm 0.3$ & $\mathbf{1.7 \pm 0.3}$ & $46.4$ \\
 & \textit{DT.} (h) & $1.3 \pm 0.3$ & $\mathbf{1.1 \pm 0.4}$ & $1.3 \pm 0.2$ & $1.3 \pm 0.3$ & $1.3 \pm 0.3$ & $0.8$ \\
 & \textit{Vf.} & $12.1 \pm 3.8$ & $\mathbf{7.6 \pm 2.5}$ & $8.5 \pm 2.9$ & $12.1 \pm 3.8$ & $12.1 \pm 3.8$ & $0.3$ \\
 & \textit{RT.} (min) & $8.8 \pm 3.9$ & $\mathbf{0.3 \pm 0.0}$ & $1.3 \pm 0.1$ & $3.5 \pm 0.3$ & $3.4 \pm 1.2$ & $-$ \\
 & \textit{Mem.} (GB) & $8.0 \pm 0.0$ & $8.1 \pm 0.2$ & $7.7 \pm 0.1$ & $8.0 \pm 0.1$ & $\mathbf{7.3 \pm 0.1}$ & $-$ \\
 & \textit{VPD (\(10^{-2}\))} & $0.5 \pm 0.2$ & $0.7 \pm 0.2$ & $0.8 \pm 0.3$ & $0.5 \pm 0.2$ & $\mathbf{0.5 \pm 0.2}$ & $0.0$ \\
 & \textit{DARD (\(10^{-2}\))} & $2.7 \pm 0.7$ & $2.5 \pm 0.2$ & $\mathbf{2.0 \pm 0.5}$ & $2.8 \pm 0.7$ & $2.8 \pm 0.7$ & $0.1$ \\
 & \textit{ATM (\(10^{-2}\))} & $8.2 \pm 0.6$ & $\mathbf{7.8 \pm 0.5}$ & $8.1 \pm 1.1$ & $8.2 \pm 0.6$ & $8.1 \pm 0.6$ & $0.0$ \\
\midrule

Shanghai & \textit{$\Delta r$} & $\mathbf{1.9 \pm 0.0}$ & $21.3 \pm 0.3$ & $2.0 \pm 0.0$ & $\mathbf{1.9 \pm 0.0}$ & $1.9 \pm 0.0$ & $0.0$ \\
 & \textit{$r_g$} & $\mathbf{1.8 \pm 0.0}$ & $14.5 \pm 0.0$ & $1.8 \pm 0.0$ & $\mathbf{1.8 \pm 0.0}$ & $1.8 \pm 0.0$ & $0.0$ \\
 & \textit{TD.} (min) & $5.4 \pm 0.0$ & $23.5 \pm 0.3$ & $5.4 \pm 0.0$ & $5.2 \pm 0.0$ & $\mathbf{3.9 \pm 0.0}$ & $-$ \\
 & \textit{DT.} (h) & $2.4 \pm 0.4$ & $2.5 \pm 0.3$ & $\mathbf{2.4 \pm 0.4}$ & $2.4 \pm 0.4$ & $2.4 \pm 0.4$ & $-$ \\
 & \textit{Vf.} & $18.7 \pm 2.4$ & $20.5 \pm 1.2$ & $18.8 \pm 2.4$ & $\mathbf{18.5 \pm 2.4}$ & $18.7 \pm 2.4$ & $0.2$ \\
 & \textit{RT.} (min) & $5.0 \pm 0.4$ & $\mathbf{0.8 \pm 0.0}$ & $1.5 \pm 0.1$ & $4.4 \pm 0.6$ & $2.9 \pm 0.4$ & $-$ \\
 & \textit{Mem.} (GB) & $8.2 \pm 0.4$ & $10.2 \pm 0.4$ & $\mathbf{5.7 \pm 0.4}$ & $8.3 \pm 0.4$ & $8.3 \pm 0.3$ & $-$ \\
 & \textit{Degree} & $1.32 \pm 0.02$ & $2.47 \pm 0.01$ & $\mathbf{1.31 \pm 0.01}$ & $-$ & $1.32 \pm 0.02$ & $17.83$ \\
 & \textit{Clustering coeff.} & $\mathbf{0.09 \pm 0.00}$ & $0.47 \pm 0.00$ & $0.09 \pm 0.00$ & $-$ & $0.09 \pm 0.00$ & $0.00$ \\
 & \textit{Edge persistence} & $0.01 \pm 0.00$ & $\mathbf{0.01 \pm 0.00}$ & $0.01 \pm 0.00$ & $-$ & $0.01 \pm 0.00$ & $0.00$ \\
 & \textit{Topological overlap} & $\mathbf{0.04 \pm 0.00}$ & $0.30 \pm 0.00$ & $0.04 \pm 0.00$ & $-$ & $0.04 \pm 0.00$ & $0.00$ \\
\midrule

YJMOB & \textit{$\Delta r$} & $\mathbf{4.3 \pm 0.0}$ & $29.4 \pm 0.1$ & $\mathbf{4.3 \pm 0.0}$ & $4.4 \pm 0.0$ & $4.3 \pm 0.0$ & $0.0$ \\
 & \textit{$r_g$} & $\mathbf{9.2 \pm 0.0}$ & $14.0 \pm 0.0$ & $\mathbf{9.2 \pm 0.0}$ & $9.3 \pm 0.0$ & $9.2 \pm 0.0$ & $0.1$ \\
 & \textit{TD.} (min) & $\mathbf{2.0 \pm 0.0}$ & $29.7 \pm 0.1$ & $\mathbf{2.0 \pm 0.0}$ & $2.0 \pm 0.0$ & $2.5 \pm 0.0$ & $-$ \\
 & \textit{DT.} (h) & $1.9 \pm 0.3$ & $\mathbf{1.7 \pm 0.0}$ & $1.9 \pm 0.3$ & $2.0 \pm 0.3$ & $2.0 \pm 0.3$ & $0.0$ \\
 & \textit{Vf.} & $\mathbf{69.9 \pm 19.4}$ & $109.2 \pm 0.6$ & $70.0 \pm 20.2$ & $73.6 \pm 21.7$ & $70.2 \pm 19.3$ & $1.2$ \\
 & \textit{RT.} (min) & $15.2 \pm 0.6$ & $15.0 \pm 0.2$ & $11.4 \pm 0.1$ & $15.1 \pm 0.8$ & $\mathbf{9.1 \pm 1.8}$ & $-$ \\
 & \textit{Mem.} (GB) & $57.1 \pm 0.8$ & $\mathbf{16.8 \pm 0.0}$ & $56.5 \pm 0.8$ & $57.7 \pm 0.6$ & $56.9 \pm 0.6$ & $-$ \\
 & \textit{Degree} & $1.61 \pm 0.00$ & $2.62 \pm 0.02$ & $1.62 \pm 0.01$ & $-$ & $\mathbf{1.61 \pm 0.01}$ & $16.27$ \\
 & \textit{Clustering coeff.} & $0.09 \pm 0.00$ & $0.29 \pm 0.00$ & $\mathbf{0.09 \pm 0.00}$ & $-$ & $0.09 \pm 0.00$ & $0.00$ \\
 & \textit{Edge persistence} & $0.00 \pm 0.00$ & $\mathbf{0.00 \pm 0.00}$ & $0.00 \pm 0.00$ & $-$ & $0.00 \pm 0.00$ & $0.00$ \\
 & \textit{Topological overlap} & $0.04 \pm 0.00$ & $0.17 \pm 0.00$ & $\mathbf{0.04 \pm 0.00}$ & $-$ & $0.04 \pm 0.00$ & $0.00$ \\
\midrule

YJMOB disaster & \textit{$\Delta r$} & $7.0 \pm 0.0$ & $25.8 \pm 0.3$ & $7.0 \pm 0.0$ & $\mathbf{6.9 \pm 0.0}$ & $7.0 \pm 0.0$ & $0.2$ \\
 & \textit{$r_g$} & $10.5 \pm 0.1$ & $12.1 \pm 0.0$ & $10.5 \pm 0.1$ & $\mathbf{10.4 \pm 0.1}$ & $10.5 \pm 0.1$ & $0.2$ \\
 & \textit{TD.} (min) & $4.7 \pm 0.0$ & $25.5 \pm 0.3$ & $4.7 \pm 0.0$ & $\mathbf{4.6 \pm 0.0}$ & $5.8 \pm 0.0$ & $-$ \\
 & \textit{DT.} (h) & $2.7 \pm 0.1$ & $\mathbf{2.3 \pm 0.0}$ & $2.7 \pm 0.1$ & $2.7 \pm 0.1$ & $2.7 \pm 0.1$ & $0.1$ \\
 & \textit{Vf.} & $108.4 \pm 6.6$ & $\mathbf{93.9 \pm 4.2}$ & $106.3 \pm 9.6$ & $103.2 \pm 11.9$ & $105.8 \pm 10.0$ & $3.2$ \\
 & \textit{RT.} (min) & $7.5 \pm 1.1$ & $4.3 \pm 0.8$ & $\mathbf{3.4 \pm 0.1}$ & $5.2 \pm 1.2$ & $3.9 \pm 0.2$ & $-$ \\
 & \textit{Mem.} (GB) & $8.2 \pm 0.3$ & $\mathbf{5.2 \pm 0.1}$ & $5.5 \pm 0.3$ & $8.0 \pm 0.3$ & $8.0 \pm 0.2$ & $-$ \\
 & \textit{Degree} & $1.75 \pm 0.01$ & $2.62 \pm 0.01$ & $1.76 \pm 0.01$ & $-$ & $\mathbf{1.75 \pm 0.01}$ & $20.41$ \\
 & \textit{Clustering coeff.} & $0.10 \pm 0.00$ & $0.29 \pm 0.00$ & $\mathbf{0.10 \pm 0.00}$ & $-$ & $0.10 \pm 0.00$ & $0.00$ \\
 & \textit{Edge persistence} & $0.00 \pm 0.00$ & $\mathbf{0.00 \pm 0.00}$ & $0.00 \pm 0.00$ & $-$ & $0.00 \pm 0.00$ & $0.00$ \\
 & \textit{Topological overlap} & $0.05 \pm 0.00$ & $0.17 \pm 0.00$ & $0.05 \pm 0.00$ & $-$ & $\mathbf{0.05 \pm 0.00}$ & $0.00$ \\
\bottomrule
\end{tabular}
\end{table*}

\section{Default Profile Schema}
\label{app:profile_schema}

The default \simname{} profile generator samples the following attributes:

\begin{itemize}[noitemsep, topsep=2pt, leftmargin=*]
\item \textbf{Gender.} Gender is assigned uniformly at random by default. The distribution can be replaced by user-provided demographic data.

\item \textbf{Age.} Age is sampled from a configurable Beta distribution \cite{gupta2004handbook}, with a default range from 16 to 80 years. Distribution parameters can be manually specified or calibrated with LLM assistance to approximate the demographic profile of the target city.

\item \textbf{Education level.} Education is sampled from a multinomial distribution over five categories: no diploma, secondary or less, vocational/technical, bachelor's degree, and master's degree or above.

\item \textbf{Health level.} Health is sampled from a multinomial distribution over a five-point Likert scale.

\item \textbf{Household composition.} Household structure is sampled from a multinomial distribution over seven categories: shared household, couple with children, couple without children, other family member, single parent, living with parents, and living alone.

\item \textbf{Occupation.} Occupation is sampled from high-level ILOSTAT job classifications \cite{ilo2026ilostat}. Occupation probabilities can be inferred from the target city's spatial structure, including the distribution of points of interest and building categories.

\item \textbf{Transport resources.} Car and bicycle availability are estimated from the agent profile. To avoid per-agent inference at large scale, \simname{} can cluster similar profiles and assign transport-resource probabilities at the cluster level using either LLM-assisted calibration or a fine-tuned ModernBERT model \cite{modernBERT}.

\item \textbf{Home and work locations.} Home and work locations are sampled from spatial heuristics based on building density, POI density, and home--work distance distributions. Building and POI information is obtained from Overture Maps \cite{overturemaps2026}.

\end{itemize}

\section{Validation Metrics}
\label{app:metrics_table}

Table~\ref{tab:metrics_available} summarizes the full set of metrics and empirical laws supported by the \simname{} validation dashboard.

\section{Ablation Study}
\label{appx:ablation}

This section evaluates the simulator across four datasets (Table \ref{tab:combined_results_vertical_ablation}), assessing realism via mobility metrics and ablation studies. Note that ablating modules can artificially inflate specific metrics; for instance, deactivating the transport mode superficially improves transport duration while obscuring multi-kilometer deviations in jump lengths. For each dataset, we evenly divide the empirical population: one half configures the simulation's population size, while the other serves as the evaluation target. The Ref. column reports an empirical-to-empirical baseline computed between these two halves, estimating expected sampling variation. Overall, \simname{} successfully reproduces real-world characteristics—often achieving agreement comparable to the Ref. baseline—with the full architecture providing the best balance across all mobility metrics.

\section{Validation Dashboard}\label{app:valDash}

\begin{figure}[!htbp]
    \centering

    \begin{subfigure}{0.98\linewidth}
        \centering
        \includegraphics[width=\linewidth]{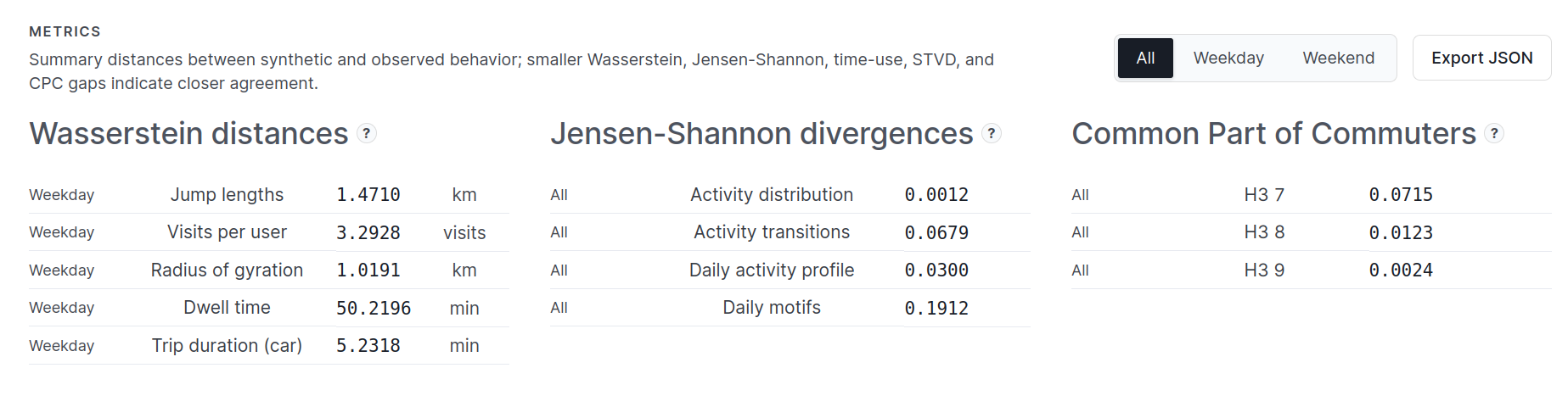}
        \caption{Comparison Metrics}
        \label{fig:comparision_metrics}
    \end{subfigure}

    \begin{subfigure}{0.48\linewidth}
        \centering
        \includegraphics[width=\linewidth]{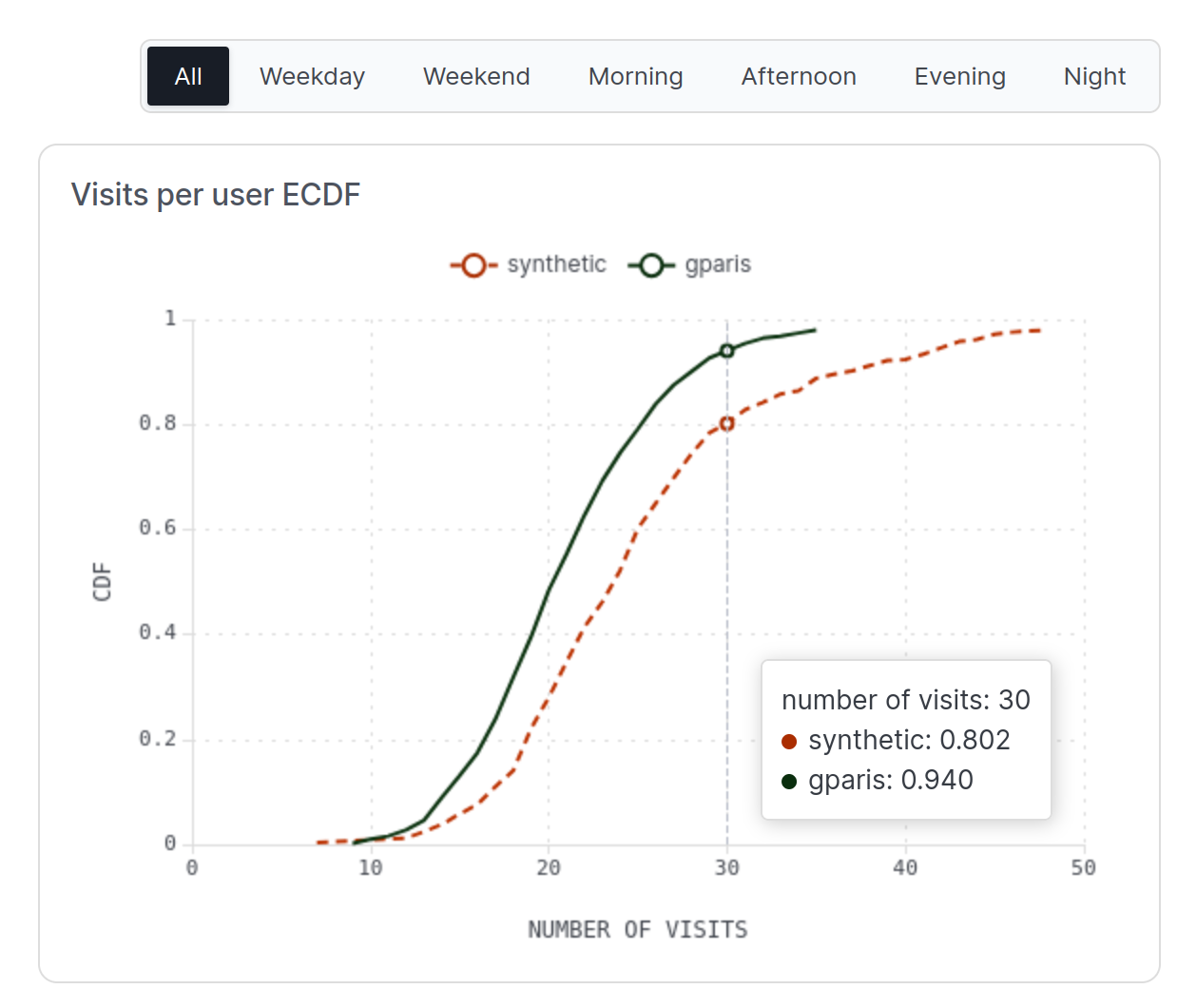}
        \caption{Distributions}
        \label{fig:sub_left}
    \end{subfigure}\hfill
    \begin{subfigure}{0.48\linewidth}
        \centering
        \includegraphics[width=\linewidth]{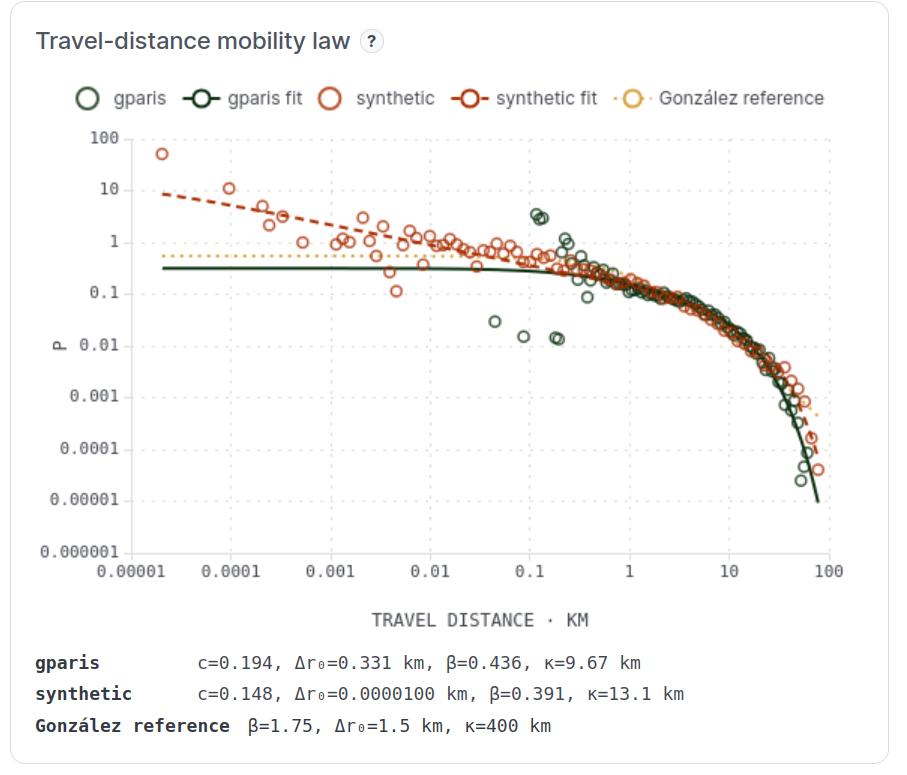}
        \caption{Mobility Laws}
        \label{fig:mobility_law}
    \end{subfigure}

    \begin{subfigure}{0.48\linewidth}
        \centering
    \includegraphics[width=\linewidth]{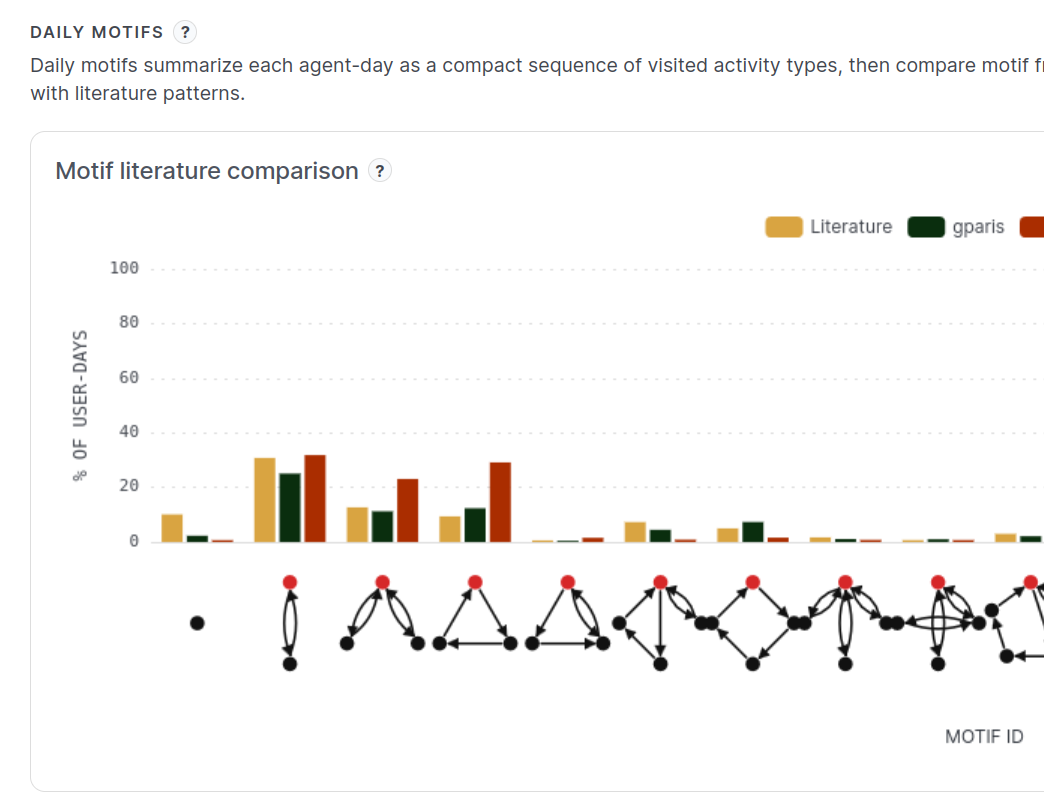}
        \caption{Motifs}
        \label{fig:motif_ui}
    \end{subfigure}\hfill
    \begin{subfigure}{0.48\linewidth}
        \centering
        \includegraphics[width=\linewidth]{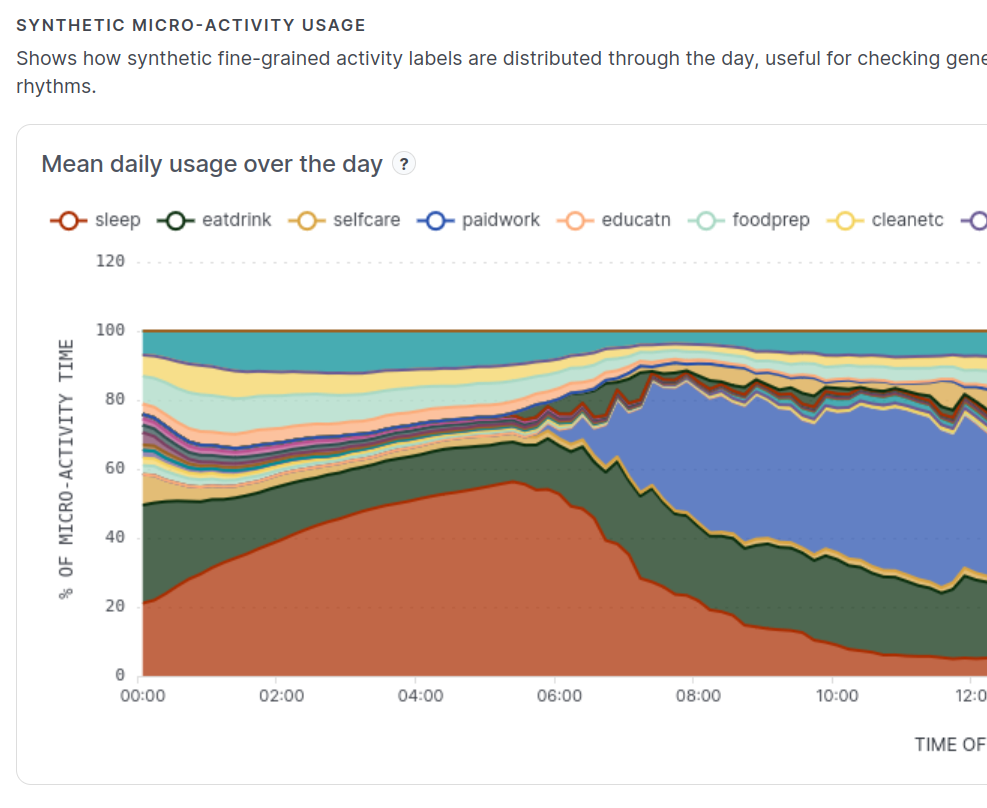}
        \caption{Time-Use}
        \label{fig:time_use}
    \end{subfigure}

    \caption{\simname{} Validation dashboard}
    \label{fig:validation_ui}
\end{figure}

Fig. \ref{fig:validation_ui} illustrates the platform's validation dashboard.

\end{document}